\pdfoutput=1
\documentclass[11pt]{article}
\usepackage{acl}

\usepackage{times}
\usepackage{latexsym}
\usepackage[T1]{fontenc}
\usepackage[utf8]{inputenc}
\usepackage{microtype}
\usepackage{inconsolata}
\usepackage{graphicx}

\usepackage{tabularx}
\usepackage{nth}
\usepackage{booktabs}
\usepackage{array}
\usepackage{amsmath}
\usepackage{multirow}
\usepackage{subcaption}
\usepackage{todonotes}

\title{Can Model Uncertainty Function as a Proxy \\for Multiple-Choice Question Item Difficulty?}

\author{Leonidas Zotos \\
  University of Groningen\\
  \texttt{l.zotos@rug.nl} \\\And
  Hedderik van Rijn \\
  University of Groningen \\
  \texttt{d.h.van.rijn@rug.nl} \\ \And
  Malvina Nissim \\
  University of Groningen \\
  \texttt{m.nissim@rug.nl}}
  
\begin{document}
\maketitle
\begin{abstract}

Estimating the difficulty of multiple-choice questions would be great help for educators who must spend substantial time creating and piloting stimuli for their tests, and for learners who want to practice. Supervised approaches to difficulty estimation have yielded to date mixed results. In this contribution we leverage an aspect of generative large models which might be seen as a weakness when answering questions, namely their \textit{uncertainty}. Specifically, we exploit model uncertainty towards exploring correlations between two different metrics of uncertainty, and the actual student response distribution. While we observe some present but weak correlations, we also discover that the models' behaviour is different in the case of correct vs wrong answers, and that correlations differ substantially according to the  different question types which are included in our fine-grained, previously unused dataset of 451 questions from a Biopsychology course. In discussing our findings, we also suggest potential avenues to further leverage model uncertainty as an additional proxy for item difficulty.

\end{abstract}

\section{Introduction}
Designing high-quality assessment methods, such as multiple-choice questions (MCQs), is both time-consuming and expensive, especially when field-testing is involved. This proves to be challenging not only due to the creation of the examination material, but also to estimating item difficulty. While it might be sensible for an exam to contain an appropriate balance of MCQs of low/middle/high difficulty, estimating such difficulty (measured by the proportion of respondents choosing the correct answer) is a complex task.

Traditionally, difficulty is estimated using expert intuitions in the creation of the stimuli, alongside pilot tests involving human subjects. Both aspects are rather time and energy consuming, so it is not surprising that especially with the massive recent advances in automatic language processing, there have been multiple efforts to address this task automatically \citep{xue-etal-2020-predicting,ni2021deepqr,rogoz2024unibucllm}. As evidence of this, the shared task organised within the latest edition of the \textit{Building Educational Applications with Natural Language Processing} workshop \citep{sharedtaskBEA2024} revolved specifically around MCQ difficulty estimation, and attracted the participation of more than fifteen teams.

\begin{figure}[t!]
    \centering
    \includegraphics[width=1\linewidth]{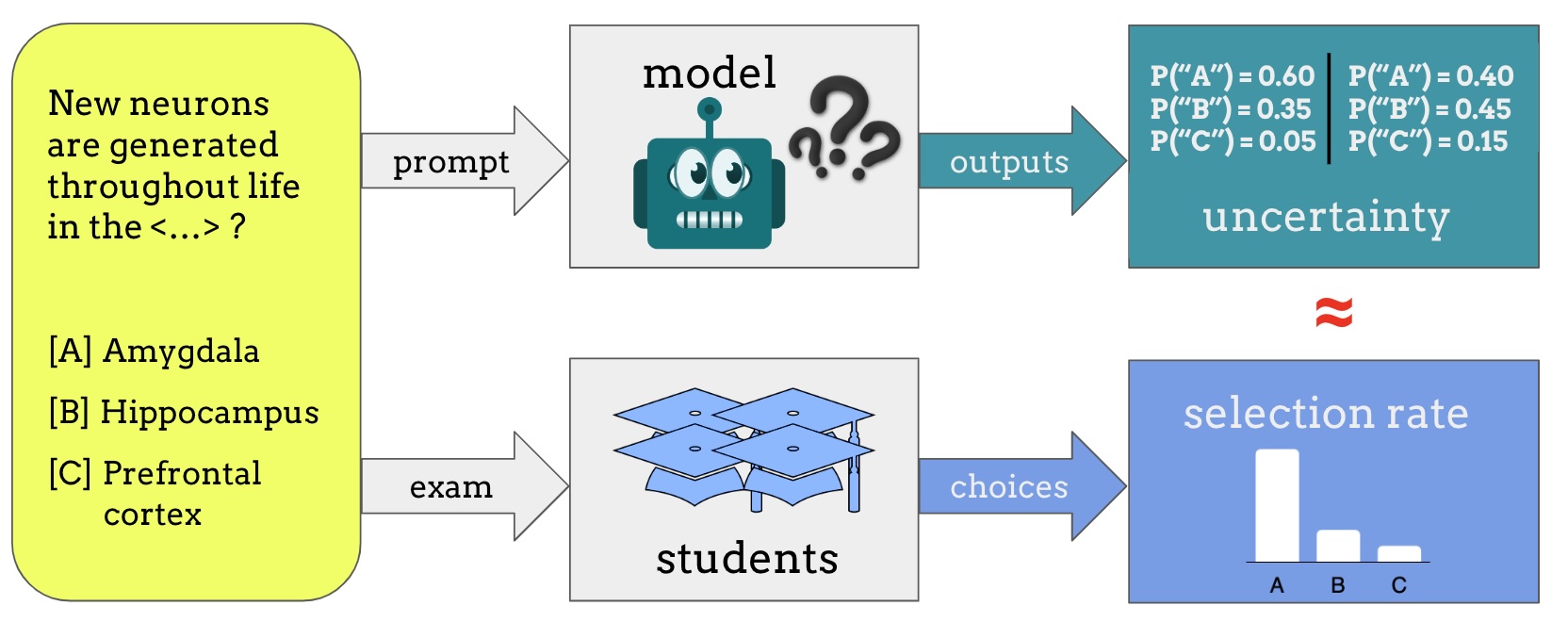}
    \caption{Exploring the potential of model uncertainty as a proxy for student selection rate in Multiple-Choice Questions. Model uncertainty is operationalised using 1st Token Probability and Choice Order Sensitivity.}
    \label{fig:main_flowchart}
\end{figure}

More in general, MCQs are all but unexplored in the modern language technology landscape. One of the main areas in which they have gathered attention is that of the evaluation of various aspects of Large Language Models (LLMs): MCQ datasets are used to assess various abilities of LLMs related to facts, knowledge, and factuality \citep[e.g.,]{srivastava2023beyond,NEURIPS2023-CMExam,lin-etal-2022-truthfulqa} for example, or reasoning strategies \citep{wang2022pinto}. In addition to mixed results in terms of accuracy, recent work has shown that LLMs answers to MCQs are particularly susceptible to the order in which the options are presented, with the first option often being favoured significantly \citep{wang2023large}, and to the prompt's formulation \cite{singhal2023towards}, so that the effectiveness of using MCQs for evaluating LLMs is under question \citep{li2024can}.

We do bring such observations into the present work, but rather than looking at MCQs as a tool to evaluate the abilities of LLMs, \textit{we look at the feasibility of using LLMs as a tool to help with MCQ difficulty estimation}. To this end, we take a novel angle compared to previous difficulty estimation studies where models are trained to predict item difficulty \citep{xue-etal-2020-predicting,ni2021deepqr,rogoz2024unibucllm}, and instead leverage \textit{model uncertainty}. Specifically, we test whether and how different uncertainty metrics, and different LLMs, can be exploited towards the prediction of MCQs' difficulty. In practice, we opt for a setting where we can also unpack the problem from a qualitative viewpoint of the questions themselves. This translates into choosing a smaller but fine-grained dataset of manually crafted MCQs which has not been published before (hence LLMs have not been exposed to this dataset in training), and which is equipped with aggregated student choices. This allows us to run a more in-depth analysis of how model choices correlate with student choices across the subtypes of questions included in the dataset. 

Our results suggest that model uncertainty can serve as proxy for MCQ difficulty to some extent, and that the type of MCQ question, and whether it is answered correctly or not, are relevant factors.

\paragraph{Our Contribution}
In contrast to previous work, we explore a fundamentally different approach to MCQ item difficulty estimation. Specifically, we explore whether model uncertainty signals show correlation with student performance, with a special focus on choice order sensitivity, a model uncertainty metric unique to MCQs. Rather than testing LLMs' capabilities using very large MCQ datasets as monolithic collections, we study students' and LLMs' behaviours at a finer-grained level, finding different degrees of correlations according to different aspects of the questions. 
Viewing LLMs as "simulated classroom environments" as we suggest in this work, paired with advancements in model interpretability metrics, contributes to devising novel approaches to the use of language technology for educational applications.\footnote{All code is publicly available:\newline \url{https://github.com/LeonidasZotos/Uncertainty-as-Proxy-for-Difficulty}.}

\section{Related Work}
Our work is positioned at the intersection of two research areas concerned with MCQs. The first is focused on the task of \textit{MCQ item difficulty estimation} using computational methods, the second focuses on \textit{understanding how LLMs answer MCQs}.

\subsection{Item Difficulty Estimation}
Previous approaches to MCQ item difficulty estimation have often focused on the training of models to complete the task. Overall, these approaches have seen limited success, showcasing the challenging nature of the task, as it has also been observed in the recent Building Educational Applications 2024 Shared Task \citep{sharedtaskBEA2024}. Here, the difficulty estimation task was cast as a regression problem, and the best team out of the seventeen that participated in the shared task performed only marginally better than a simple baseline. In contrast to the present work, the shared task focused on highly homogeneous data which did not include student selection rates for all choices, but rather only for the correct answer.

A common approach to predicting item difficulty (and also response time) is using supervision, either training a model from scratch or fine-tuning a pre-trained model. \citet{xue-etal-2020-predicting}, attempt to predict item difficulty and response time by training an ELMo model \citep{peters-etal-2018-deep}, leveraging $18,000$ MCQs from a high-stakes medical exam database. On the same dataset, \citet{yaneva-etal-2021-using} used unsupervised clustering techniques to observe that nuanced features, such as the number of ambiguous medical terms, helped explain response process complexity beyond superficial item characteristics like word count.\footnote{At time of writing, the proceedings of the BEA shared task on item difficulty estimation are not yet available, so that we do not know specifically which approaches were tested; it is however known that simple baselines were overall only marginally beaten \cite{sharedtaskBEA2024}.}

Relying on a large dataset and a variety of features, "DeepQR" is a supervised model that predicts the quality of MCQs based on human-provided quality ratings \citep{ni2021deepqr}. To strengthen the performance of the system, the authors supplement the MCQ with additional features by employing an automatic extraction of explicitly-defined features (e.g., readability, length of components) and semantic features. While performing better than simpler models the authors compare DeepQR to, this system relies on possibly less ecologically labelled data: judgements on item difficulty are collected by explicitly asking participants to provide a difficulty rating; in the present work we instead derive difficulty from the observed distribution of answers to the questions.

Most recently, \citet{raina2024question} have adapted reading comprehension systems for difficulty ranking of MCQs.  By comparing this strategy to zero-shot prompting, the authors conclude that the latter is superior in assessing item difficulty. Although their work bears some resemblance with our study, we shift the focus towards factual knowledge questions, instead of reading comprehension.

Lastly, a study by \citet{loginova2021} is perhaps most similar to our research. The authors explore using model certainty as a proxy for difficulty. However, their work involves presenting questions in pairs, with the task of selecting the more difficult one. Additionally, their focus is on language comprehension, which differs from our emphasis on factual knowledge assessment. Furthermore, while their focus is on Encoder-Only models, ours is on Decoder-Only models, which incorporate greater amounts of factual knowledge as a byproduct of their language modeling objective \cite{zhao2023survey}. 

The present work fundamentally differs from these previous studies by exploring the usefulness of model certainty for item difficulty estimation.

\subsection{LLM Bias in the MCQ Task}

In recent literature, Large Language Models undertaking the MCQ task have consistently been found to have a bias towards specific options through a combination of positional, e.g., preferring the 3rd choice \citep{pezeshkpour2023large} or token-preferential effects, e.g., bias towards generating "A"  \citep{zheng2024large}. At the same time, it has also been observed that this effect correlates with model correctness: for items where the choice order has little effect, models are more likely to be correct \cite{pezeshkpour2023large}. This is is line with the work of \citet{plaut2024softmax} which finds that lower maximum softmax probabilities correlate with incorrect answers to MCQs.

Two main methods have been used to mitigate the choice order effect. The first strategy, named "PriDe", is a method that debiases a model's predictions by estimating and correcting for the model's prior bias towards certain option IDs (e.g., ID of "A"); this is done by permuting option contents on a subset of test samples and then applying the estimated prior to debias the remaining samples \citep{zheng2024large}. The second strategy, proposed by \citet{wang2024look} suggests that this effect can be simply mitigated by focusing on the textual output, rather than the 1st token probabilities. While both of these strategies highlight --- and focus on reducing --- the choice order effect, we leverage this finding and explore if the magnitude of this effect can be useful in determining the item's difficulty.

Lastly, some recent research has investigated the extent to which LLMs reflect human response biases in their responses to opinion surveys with different perturbations such as choice order, acquiescence, and opinion floating \citep{tjuatja2024llms}. Findings show that LLMs are sensitive to perturbations that do not however elicit significant changes in humans, and are overall not great proxies for human behaviour. In our work we explore their proxy potential when dealing with factual MCQs with single correct answers rather than subjective aspects. 

\section{Methodology}
Figure~\ref{fig:main_flowchart} captures the central design of our study: we explore if model uncertainty can serve as a proxy for MCQ difficulty. Difficulty is obtained through student response distribution; Section~\ref{sec:dataset} describes the dataset we use, focusing on student performance. Model uncertainty is obtained through the different metrics that we describe in Section~\ref{sec:measuring_model_uncertainty}. In Section~\ref{sec:decoder-only} we explain our  requirements and choices regarding models and prompting used in our experiments.

\subsection{Student Performance Dataset}
\label{sec:dataset}
For this study, a set of MCQs and student performance data, such as the proportion of students selecting each choice, is required. Unfortunately, such a dataset is not publicly available. Therefore, this study utilises data from the "Biopsychology" course at the Social Sciences Faculty of the University of Groningen, covering content from the classic textbook "Biological Psychology" by \citet{Kalat_2016}. 

The dataset comprises $451$ MCQs. For each question, there are three possible answer choices, and the dataset records the fraction of students who selected each of those three options. The data was collected from eight examinations with an average of $268$ examinees (Standard Deviation of $185$). Moreover, the questions were designed by the course instructor and have not been previously published, minimising the risk of data contamination. 

An important aspect of this dataset is its variety in terms of formulation types, as shown in Table~\ref{tab:question_formulations}. The questions follow four main formulations: "Fill the gap" (14.9\%), "Fill two gaps" (3.1\%), "Wh-question" (50.3\%), and "Sentence Completion" (31.7\%). For brevity, we will refer to these as question types 1 through 4 respectively.

\begin{table*}[ht]
\begin{small}
\centering
    \begin{tabular}{m{0.1\textwidth} m{0.47\textwidth}  m{0.33\textwidth}}
        \toprule
        \textbf{Type} & \textbf{Question} & \textbf{Choices} \\ 
        \midrule
        \textit{Fill the gap} & In explaining colour perception, the ... theory applies to what happens in the brain. & \textcircled{a} \textcolor{ForestGreen}{retinex}\newline \textcircled{b} trichromatic \newline \textcircled{c} opponent process \\
        \midrule
        \textit{Fill two gaps} & Homeostasis is to ... as allostasis is to ... & \textcircled{a} \textcolor{ForestGreen}{constant; variable}\newline \textcircled{b} constant; decreasing\newline \textcircled{c} variable; constant \\
    \midrule
       \textit{Wh-question} & While discussing their upcoming bio exam, John says that a neuron that has a larger number of dendritic branches can produce a stronger action potential. Mary replies that she thinks that a larger number of dendritic branches only increases the neuron’s ability to receive signals from other neurons. Which of these statements is correct?& \textcircled{a} \textcolor{ForestGreen}{Only Mary's statement is correct}\newline \textcircled{b} Only John's statement is correct\newline \textcircled{c} Both statements are correct \\ \midrule
        \textit{Sentence Completion} & The central nervous system consists of the ... & \textcircled{a} \textcolor{ForestGreen}{brain and spinal cord}\newline \textcircled{b} subcortical structures of the brain\newline \textcircled{c} the brainstem and cerebellum\\ 
        \bottomrule
    \end{tabular}
    \caption{Examples of different question formulations present in the dataset. Correct answer in 
    \textcolor{ForestGreen}{green}.}
\label{tab:question_formulations}
\end{small}
    
\end{table*}

\begin{figure}[h]
    \centering
    \includegraphics[width=.95\linewidth]{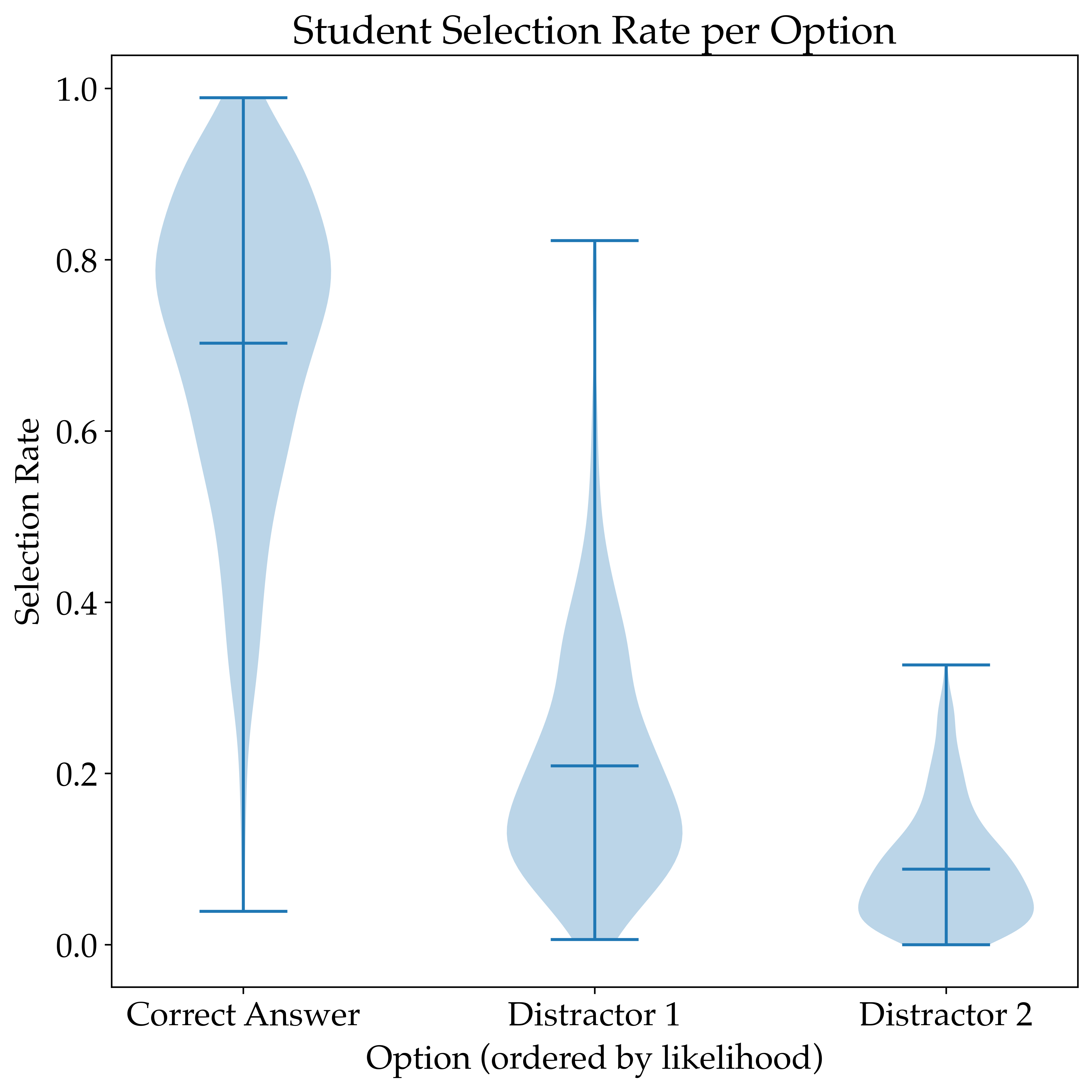}
    \caption{Average selection rate per choice. Distractors are ordered by their selection rate.}
    \label{fig:student_performance}
\end{figure}

In terms of performance, on average, 70.3\% of the students selected the correct answer. Having ordered the false choices, henceforth \textit{distractors}, by the proportion of students selecting each, distractor 2 and 3 were selected by 20.9\% and 8.8\% of students respectively, as shown in Figure~\ref{fig:student_performance}.

\subsection{Measuring Model Uncertainty} 
\label{sec:measuring_model_uncertainty}
Our study's central goal is to find whether item difficulty (measured by the proportion of students correctly answering the MCQ) correlates with model uncertainty (Figure ~\ref{fig:main_flowchart}). We use two different techniques to capture model uncertainty: \textit{1st Token Probability} and \textit{Choice Order Sensitivity}. Moreover, we calculate the entropy of the 1st token probabilities for each choice, a metric that quantifies model uncertainty overall for a given question.

\paragraph{1st Token Probability}
The first technique to measure model uncertainty is by inspecting the probability of the 1st token to be generated (e.g., probability of generating token "B"), in comparison to the probabilities of the alternatives (e.g.,  probability of generating token "A" or "C"). As the 1st token probabilities can be influenced by the order of the choices, we create six different orderings of the questions to cover all possible permutations and let the model answer each MCQ six times; we take as probability of each choice the average over all possible choice orderings.

Furthermore, as there might be multiple tokens representing the same answer (e.g., "A", " A", "a ") and models attribute higher likelihood for a specific token, the token representing each choice with the highest probability is selected. Lastly, the three extracted mean probabilities of all orderings are normalised in the range of $0-1$ such that they can be more easily compared to student response rates. 

\paragraph{Choice Order Sensitivity}
\citet{pezeshkpour2023large} observed that choice order sensitivity correlates with error rate. In other words, when LLMs consistently select a choice regardless of its position, that choice is more likely to be correct. Based on this observation and on results we obtain along the same lines\footnote{We report details in Appendix~\ref{app:order_sensitivity_results}.}, we leverage this correlation to measure model uncertainty. Specifically, for all six possible orderings, the probability of each choice being selected is measured. It is worth noting that this probability is not based on the token probabilities but rather on the eventual choice.

\paragraph{Choice Entropy}
The previously presented "1st Token Probability" and "Order Sensitivity" metrics measure the model's uncertainty \textit{per-choice} and can be evaluated as proxies for the proportion of students selecting each choice of the MCQ. To expand on these metrics and with a focus on the model's certainty for \textit{all question choices}, we consider the entropy of the 1st token probabilities of the three answer choices. Equation \ref{eq:entropy_model} describes this measure, where $H_{\text{model}}(Q)$ is the entropy for question $Q$ and $\overline{P}(Q)_c$ represents the probability of choice $c$ being generated averaged across all possible choice presentation orders. 

\begin{equation}
    \label{eq:entropy_model}
    H_{\text{model}}(Q) = -\sum_{c=0}^{2}{\overline{P}(Q)_c \cdot \ln \overline{P}(Q)_c}
\end{equation}

In this study, the equivalent measure for the student performance is considered, whereby the proportion of students selecting each choice is considered in the place of $\overline{P}(Q)_c$. 

\subsection{Choice of Models and Prompting}
\label{sec:decoder-only}
The current work focuses on Decoder-Only models, as these models are generally larger compared to Encoder-Only or Encoder-Decoder models and are considered to have incorporated greater amounts of factual knowledge as a byproduct of their language modelling objective \cite{zhao2023survey}. However, the free-text generation capability of Decoder-Only models is not optimal for MCQ answer generation. To adapt them for this task, and following the work of \citet{plaut2024softmax}, the current work also uses two prompt phrasings that instruct the model to respond with only one letter corresponding to the correct answer. Appendix~\ref{app:prompt_phrasing_results} reports the two instruction phrasings and their effect on the results. We ran this additional analysis as prompt formatting has been shown to potentially have a significant effect on the obtained results \citep{sclar2023quantifying}. Apart from Mistral-7b, where correlations are slightly stronger with phrasing~1, no substantial differences are observed; we thus report all analyses based on phrasing~1 only, as also done in \citep{plaut2024softmax}.

As described in Section~\ref{sec:measuring_model_uncertainty}, to evaluate the uncertainty of the models' answers the internal logit probabilities of the 1st token to be generated are used. As access to model internals is needed, we focus on open-sourced models of different capabilities and parameter sizes. Additionally, we constrict our choice to instruction-tuned models and use 4-bit quantisation for increased efficiency. The four models eventually used in this research are Llama3-8b-Instruct\footnote{\url{https://huggingface.co/unsloth/llama-3-8b-Instruct-bnb-4bit}}, Llama3-70b-Instruct\footnote{\url{https://huggingface.co/unsloth/llama-3-70b-Instruct-bnb-4bit}}, Yi-34b-Chat\footnote{\url{https://huggingface.co/unsloth/yi-34b-chat-bnb-4bit}}, Mistral-7b-Instruct\footnote{\url{https://huggingface.co/unsloth/mistral-7b-instruct-v0.2-bnb-4bit}}, covering a good range of sizes. For brevity, we henceforth omit the "instruct"/"chat" label when referring to the selected models.

\section{Results}
For all subsequent analyses, we use instruction phrasing 1 (see Section~\ref{sec:decoder-only} and Appendix~\ref{app:prompt_phrasing_results}). We analyse the relationship between the performance of the selected models and students through three incremental lenses, working at the per-question and per-choice levels. However, before discussing these analyses, it is useful to first examine the overall performance of the models on the dataset. Table~\ref{tab:overall_performance} summarises the proportion of correctly answered questions (based on 1st token probabilities) per question type, as defined earlier. As will be shown later, drawing a distinction between question types is imperative in studying the relationship between model certainty and student performance.

As expected, the largest model among those we tested, Llama3-70b, achieved the best performance by answering nearly 90\% of the questions correctly. For all models, questions of \text{Type~2}  ("Fill two gaps") proved to be the most challenging, which is understandable given the more complex linguistic task they present for LLMs. Moreover, since these models were pre-trained on fill-mask tasks, it is unsurprising that they performed best on question Types~1 and~4 ("Fill the gap" and "Sentence Completion," respectively), with the exception of Mistral-7b. Looking at the student performance, no substantial differences are observed between the question types, and the only question type where model performance is worse is indeed Type~2.\footnote{The limited number of examples in this category does not allow for strong statements, but we observe this behaviour.}

\begin{table}[h!]
\centering
\resizebox{\columnwidth}{!}{%
\begin{tabular}{lccccc}
\toprule
\multirow{2}{*}{\textbf{Model}} & \multicolumn{5}{c}{\textbf{Question Type}}                                         \\ \cline{2-6} 
 &
  \textbf{\begin{tabular}[c]{@{}c@{}}1\\ (n=59)\end{tabular}} &
  \textbf{\begin{tabular}[c]{@{}c@{}}2\\ (n=20)\end{tabular}} &
  \textbf{\begin{tabular}[c]{@{}c@{}}3\\ (n=227)\end{tabular}} &
  \textbf{\begin{tabular}[c]{@{}c@{}}4\\ (n=145)\end{tabular}} &
  \textbf{\begin{tabular}[c]{@{}c@{}}All\\ (n=451)\end{tabular}} \\ \midrule
Mistral-7b             & 0.797 & {0.750}  & {0.652} & 0.848 & 0.738 \\
Llama3-8b              & 0.814 & 0.600   &{0.709} & 0.834 & 0.758 \\
Yi-34b                 & 0.915 & {0.750}  & {0.793} & 0.903 & 0.843 \\
Llama3-70b             & \textbf{0.966} & {0.750}  & \textbf{0.868} & \textbf{0.938} & \textbf{0.898} \\
 \midrule
\textbf{Student Rate}  & 0.693 & \textbf{0.752} & {0.691} &  0.717 & 0.703 \\ \bottomrule
\end{tabular}
}
\caption{Model and student performance on the dataset used in this study. Model performance is based on the choices determined by the highest 1st token probability, while student performance is measured by the proportion of students selecting the correct answer choice. Best performance per column is bolded.}
\label{tab:overall_performance}
\end{table}

\subsection{Question-Level: Correlation of Entropy}
\label{sec:question_level_correlation_entropy}
We begin the analysis by examining the relationship between student performance and model certainty, as measured by the "Choice Entropy" metric defined previously. For each question, this metric quantifies how dispersed the model's probability distribution is across the three possible answer choices. Similarly, for student performance, it quantifies how divided the student population was on that particular question. 

\begin{figure}[h!]
    \centering
    \includegraphics[width=\linewidth]{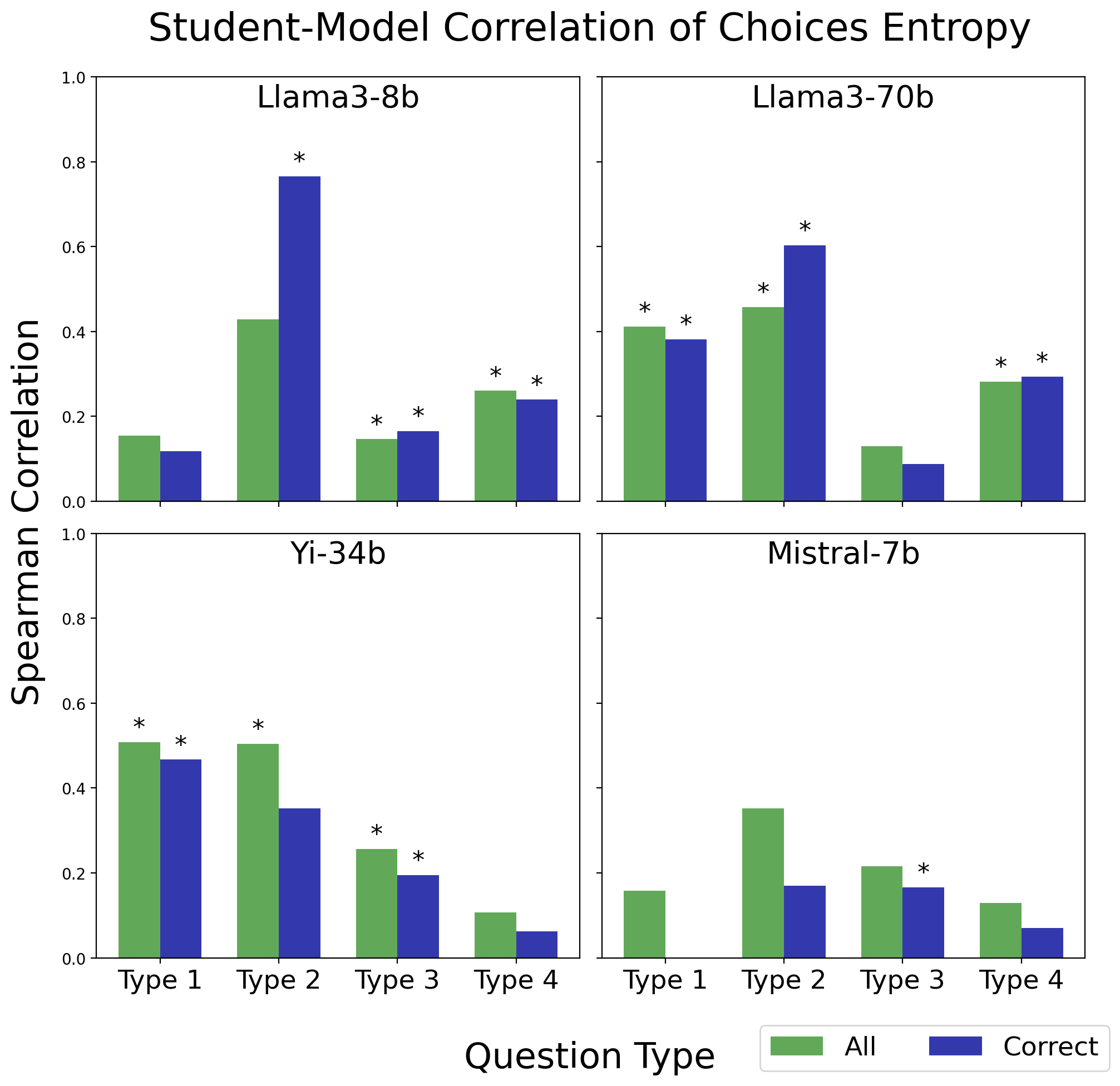}
    \caption{Spearman Correlation (implemented using the SciPy package \cite{2020SciPy-NMeth}) between student and model choice entropy. Asterisks signify significant correlation, using a significance level of $\alpha = 0.05$.}
    \label{fig:corr_entropy}
\end{figure}

Figure~\ref{fig:corr_entropy} illustrates the correlation between the choice entropy of students and models. We further categorise the questions into two subsets: all questions, and those correctly answered by each model. This additional analysis allows us to evaluate if model correctness influences how closely model certainty correlates with student selection rates.

\noindent From the figure, we can observe differences in correlation across the question types. For example, with the Llama3-8b model, the correlation for Type~2 is much stronger than for Type~3, especially when considering the subset of correctly answered questions. Mistral-7b, instead, shows weaker correlation throughout. Overall, the significant correlations found can be considered to be weak to moderate. While this analysis suggests that there is indeed some correlation between the model certainty dispersion and the student selection rates for a particular question, we need to further study it by looking into how closely the distributions of choice proportions between models and students align.

\subsection{Question-Level: Chi-Squared of Rates}
\label{sec:question_level_chi_rates}
To quantify the alignment between model certainty dispersion and the student selection rates, we perform a Chi-Squared test for each question. However, since the Chi-Squared test requires each choice to have a non-zero student proportion, we remove ten questions where at least one choice had a 0\% student selection rate.\footnote{Among the removed questions, three belong to Type~1, two to Type~3, and five to Type~4.} This results in a dataset of 441 questions for this analysis. Lastly, this analysis is conducted using both model uncertainty metrics (1st Token Probability and Choice Order Sensitivity) as described in Section~\ref{sec:measuring_model_uncertainty}. 

\begin{figure*}[ht!]
\centering
\begin{subfigure}{0.5\linewidth}
  \centering
  \includegraphics[width=\linewidth]{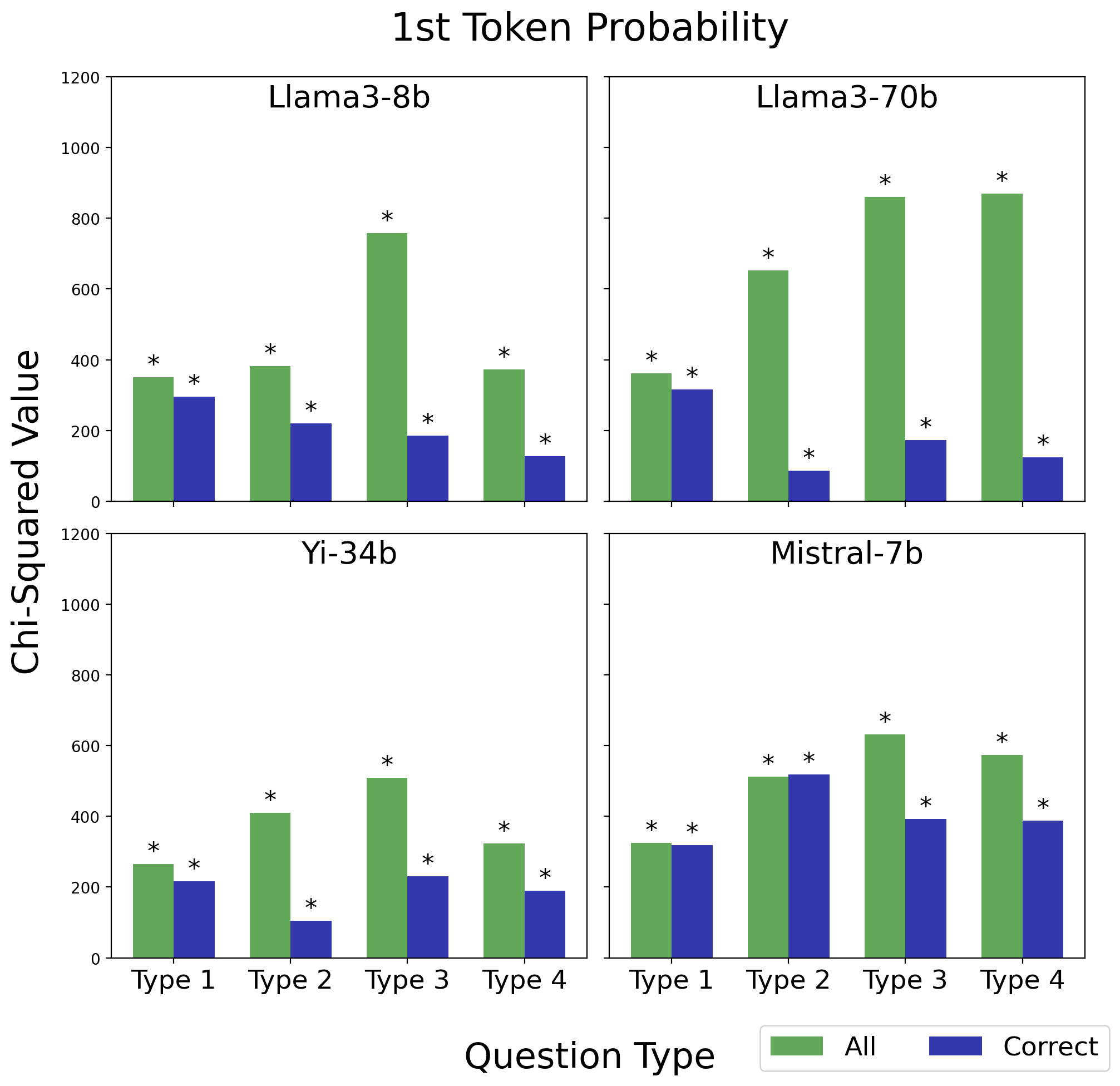}
  \label{fig:chi_squared_1st_token}
\end{subfigure}%
\begin{subfigure}{0.5\linewidth}
  \centering
  \includegraphics[width=\linewidth]{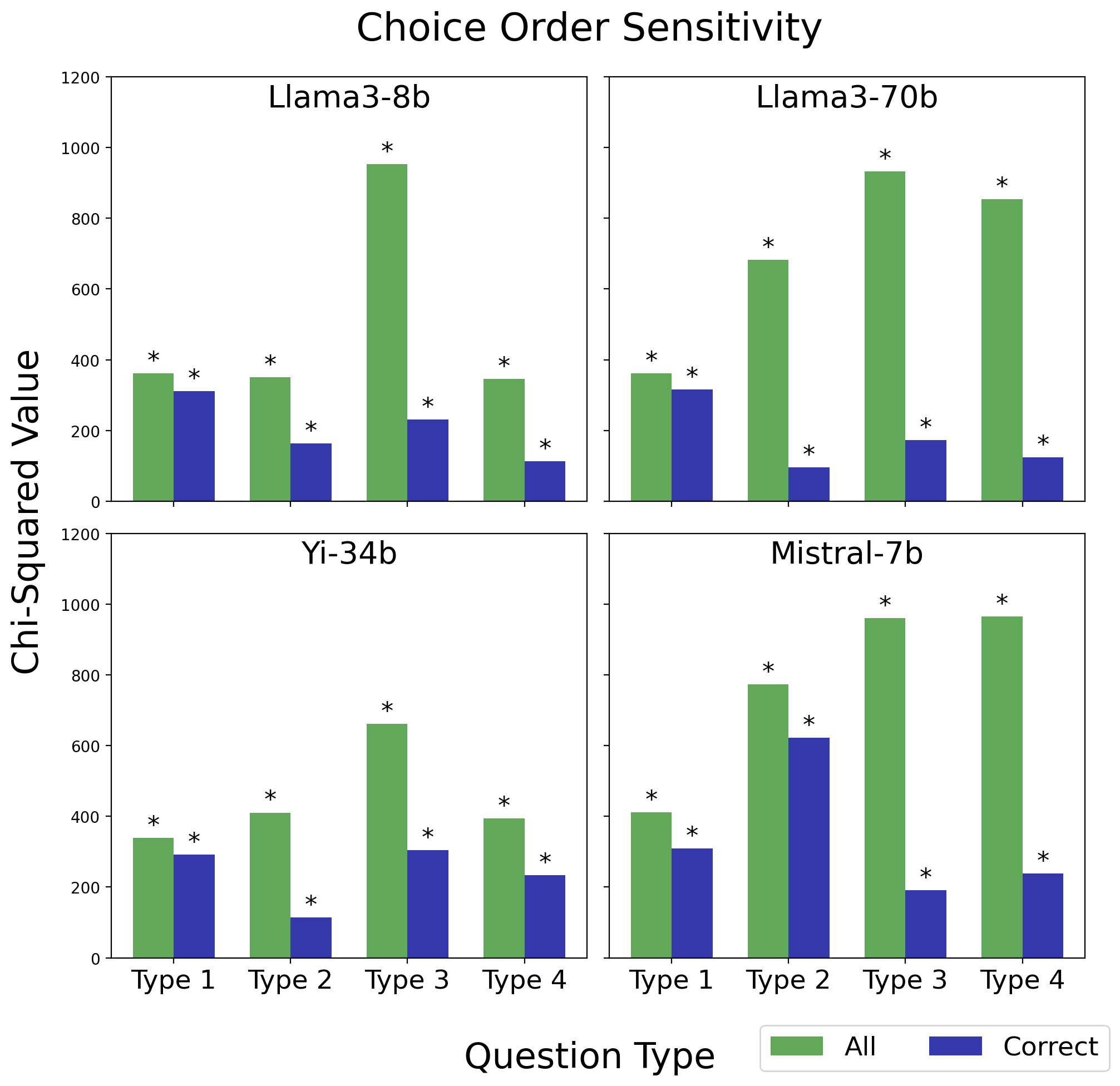}
  \label{fig:chi_squared_order}
\end{subfigure}
\caption{Average Chi-Squared value between the student proportions and model uncertainties distributions. All Chi-Squared Values indicate a statistically significant difference between the distributions, using a significance level of $a=0.05$, indicating that the student and model metrics originate from significantly different distributions.}
\label{fig:chi_squared_all}
\end{figure*}

Figure \ref{fig:chi_squared_all} illustrates the Chi-Squared value averaged for the entire set. Overall, the measurements are high, signifying that the student and model metrics originate from statistically significantly different distributions, which is expected. However, while all model distributions differ from the student distribution, certain models and question types have distributions that align more closely with student performance. Specifically, Type~1 seems to consistently have model distributions that are closer to student performance compared to Types~3 or~4 across all models, findings which are in line with the results previously presented in Section~\ref{sec:question_level_correlation_entropy}. 

It is interesting to note that focusing on the subset of correctly answered questions yields overall lower chi-squared values, particularly for the two Llama models. It is hypothesised that this difference arises from the removal of confidently incorrectly answered questions, which would have had the largest impact on the statistic. However, further analysis should be conducted to investigate this.

It is noteworthy that we observe a resemblance between the general pattern of the 1st token probabilities and the choice order sensitivities. Table~\ref{tab:corr_model_uncertainties} confirms this, showing strong correlation between the two methods of model uncertainty estimation, with the exception of LLama3-70b. This exception is not very surprising and stems from the model's insensitivity to the order bias.\footnote{Further details on this are provided in Appendix~\ref{app:order_sensitivity_results}.} 

\begin{table}[h!]
\centering
\resizebox{\columnwidth}{!}{%
\begin{tabular}{lccc}
\toprule
\multirow{2}{*}{\textbf{Model}} & \multicolumn{3}{c}{\textbf{Correlation Per Choice}} \\ 
\cmidrule(lr){2-2} \cmidrule(lr){3-3} \cmidrule(lr){4-4}
                    & \textbf{Correct Answer} & \textbf{Distractor 1} & \textbf{Distractor 2} \\ \hline
Llama3-8b & 0.979          & 0.754        & 0.743        \\
Llama3-70b & 0.633          & 0.425        & 0.530        \\
Yi-34b    & 0.934          & 0.854        & 0.874        \\
Mistral-7b & 0.896          & 0.824        & 0.814       \\\bottomrule

\end{tabular}%
}
\caption{Spearman correlation between 1st Token Probability and Choice Order Sensitivity over the complete dataset. All correlations are found to be significant, using a significance level of $a=0.05$.}
\label{tab:corr_model_uncertainties}
\end{table}

\subsection{Choice-Level: Correlation of Rates}
For a finer-grained analysis, we study the correlation between the student selection rate \textit{per choice} and the two model uncertainty metrics. 

\begin{figure*}[ht!]
\centering
\begin{subfigure}{0.5\linewidth}
  \centering
  \includegraphics[width=\linewidth]{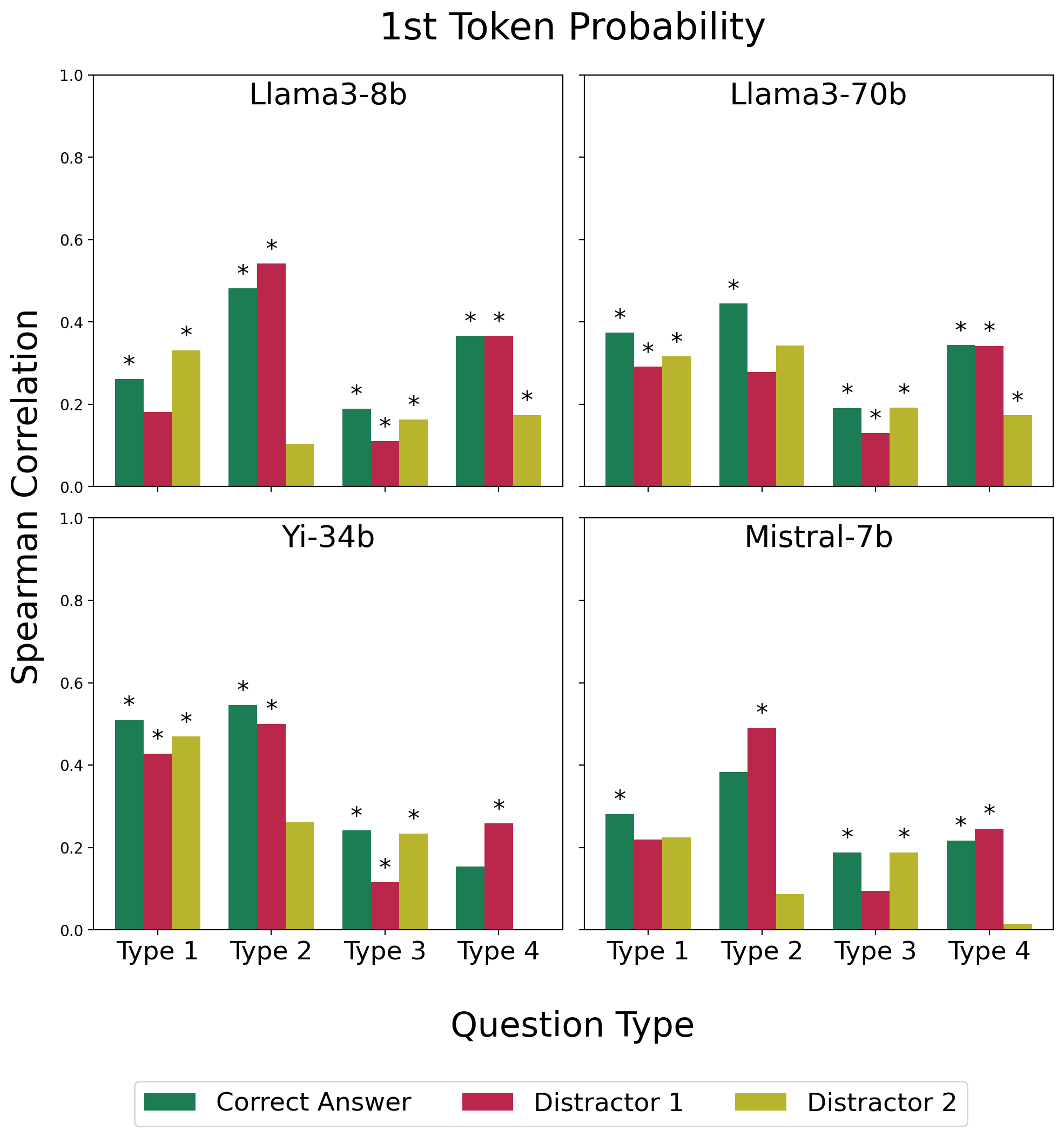}
  \label{fig:spear_1st_token_all}
\end{subfigure}%
\begin{subfigure}{0.5\linewidth}
  \centering
  \includegraphics[width=\linewidth]{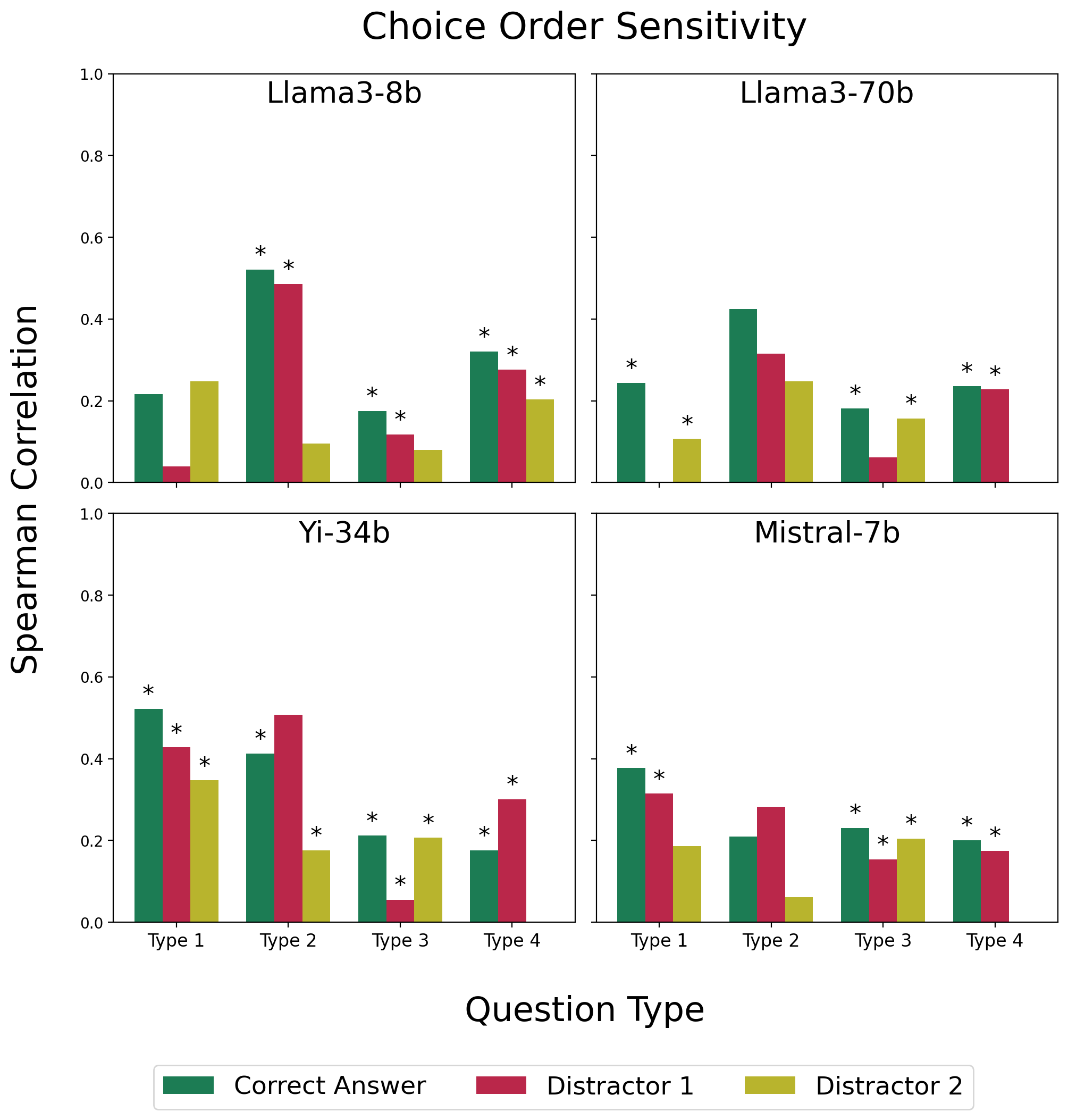}
  \label{fig:spear_order_all}
\end{subfigure}
\caption{Spearman Correlation between model uncertainty metrics and student selection rates, per choice, using the complete dataset. Asterisks signify that the correlation is significant, using a significance level of $\alpha = 0.05$.}
\label{fig:spear_all}
\end{figure*}

First, we focus on the entire dataset, containing both correctly and incorrectly answered questions, as shown in Figure~\ref{fig:spear_all}. We observe that the correlations are weak ($\rho <0.5$), but statistically significant. The findings are largely consistent with the results of previous analyses (cf. Figure~\ref{fig:corr_entropy}), with the correlation often being weakest for question Type~4, especially when compared to question Type~2.

Furthermore, in most cases, the correlation is stronger for the correct answer choice compared to distractors 1 and 2. However,  there does not seem to be consistent difference between the three choices, especially between distractors 1 and 2.

We further focus on the subset of correctly answered questions. Again, we do not observe consistent difference between the three choices. Additionally, in line with the previous analysis, the correlation for the questions of Type~2 significantly increases by only studying the subset of correctly answered questions, with the exception of the results of Mistral-7b. It is not immediately clear why Type~2 overall shows stronger correlation. We hypothesise that this is because "fill two blanks" questions are the most challenging for the tested models (cf. Table~\ref{tab:overall_performance}), so it is expected for the models to have lower certainty scores on those questions, which might be more likely to correlate with lower student selection rates \footnote{As also mentioned earlier, the limited number of examples in this category does not allow for strong statements, but we observe this behaviour.}. 

\begin{figure*}[h!]
\centering
\begin{subfigure}{0.5\linewidth}
  \centering
  \includegraphics[width=\linewidth]{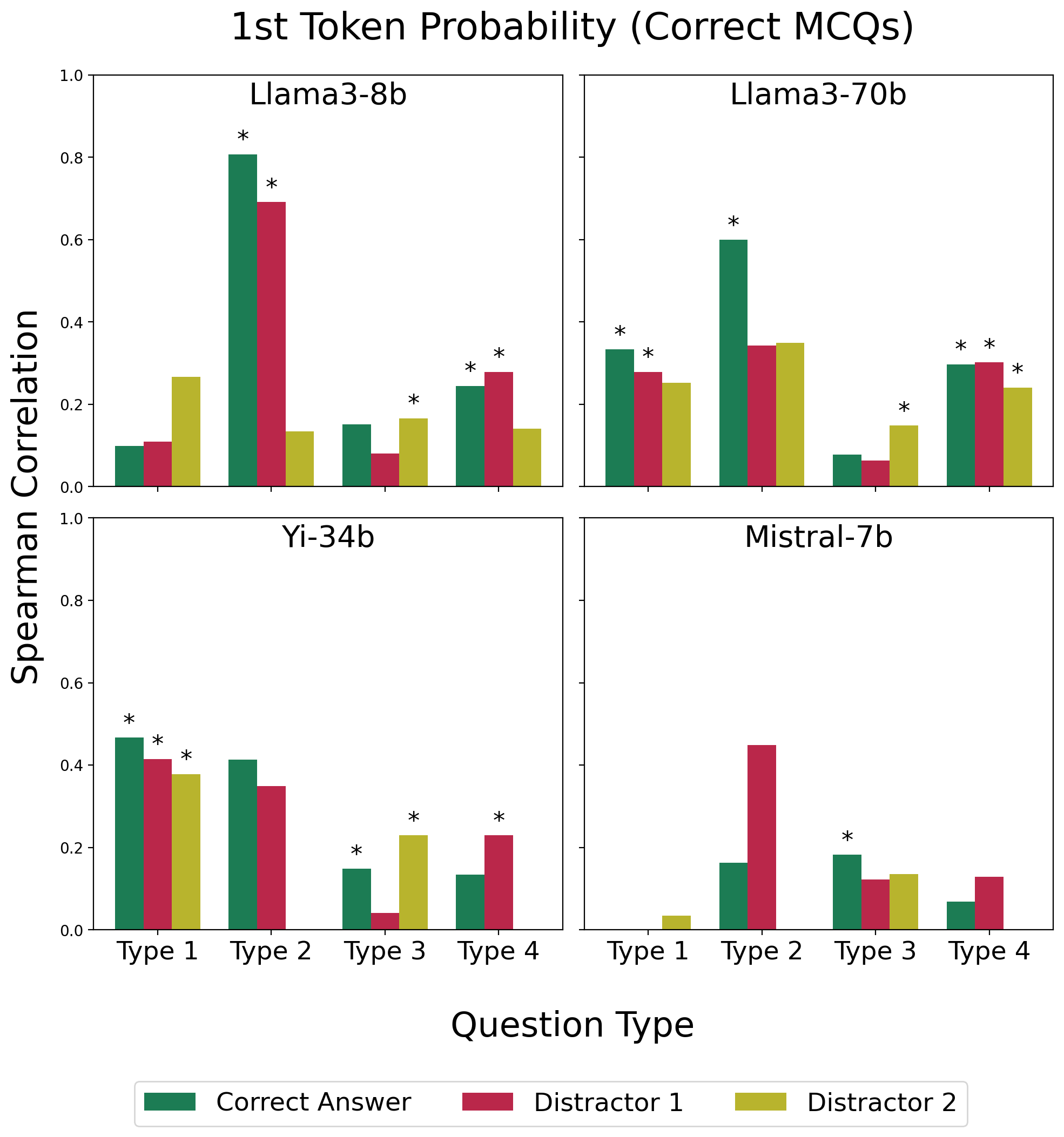}
  \label{fig:spear_1st_token_correct}
\end{subfigure}%
\begin{subfigure}{0.5\linewidth}
  \centering
  \includegraphics[width=\linewidth]{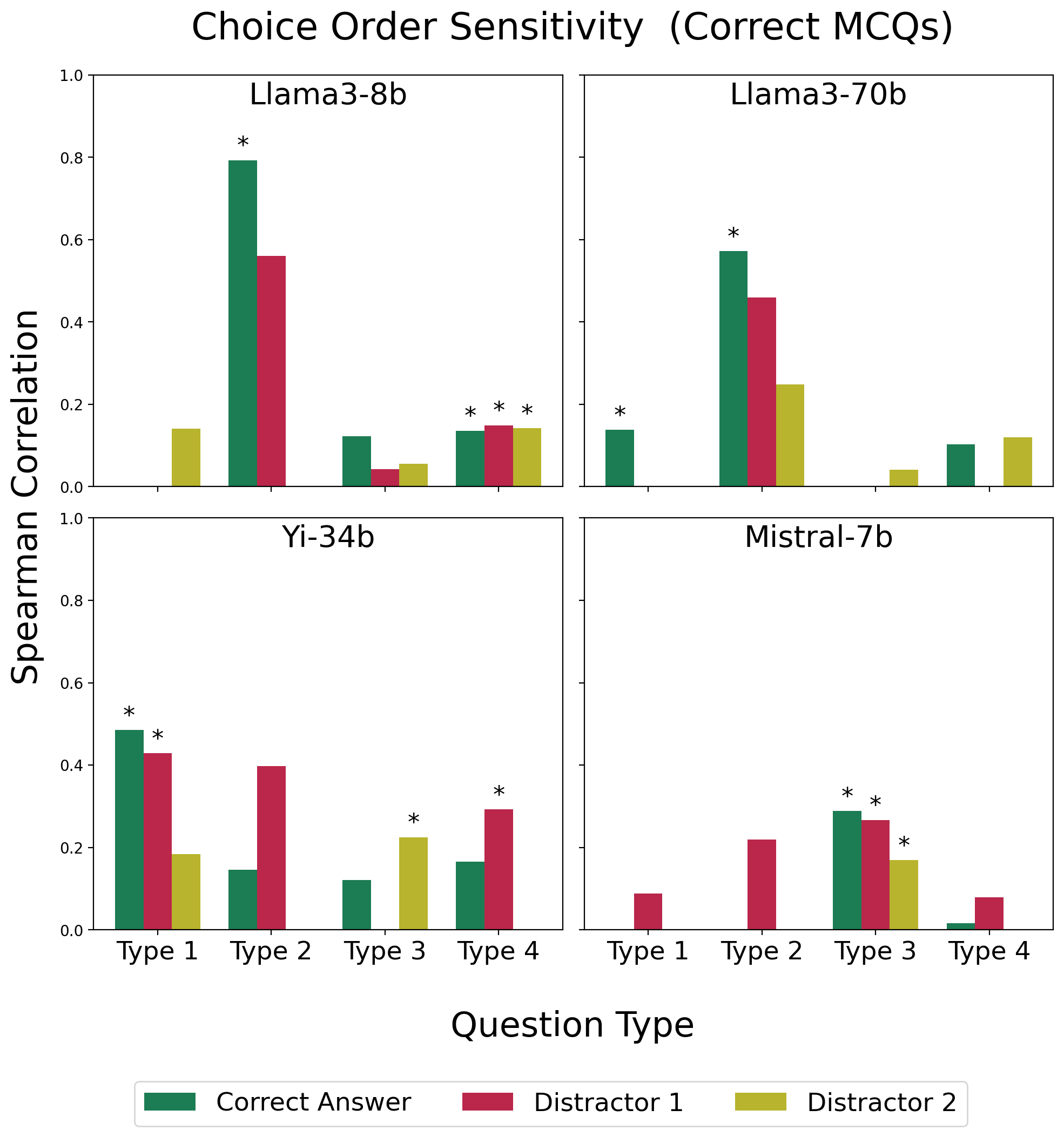}
  \label{fig:spear_order_correct}
\end{subfigure}
\caption{Spearman Correlation between model uncertainty metrics and student selection rates, per choice, using the subset of correctly answered questions. Asterisks signify that the correlation is significant, using a significance level of $\alpha = 0.05$.}
\label{fig:spear_correct}
\end{figure*}

\section{Discussion and Conclusion}
In this study, we investigated whether model uncertainty bears potential to be used for the MCQ item difficulty estimation task. Through the use of a fine-grained dataset, two model uncertainty metrics and three sets of analyses at different levels, we find that there is generally a weak correlation between model uncertainty and item difficulty. We further find that this correlation is higher for certain question types, particularly  "fill the blank" and "fill two blanks" questions. As is to be expected, we also find that the observed correlations are model-dependent, with both the weakest (Mistral-7b) and the strongest (Llama3-70b) models showing lower correlation between model uncertainty and student selection rate compared to the other models. Lastly, studying the correlation per-choice, instead of per-question, did not reveal any consistent trends.

Shifting the focus to the bigger picture, through this research, we demonstrate that it can be beneficial to consider non-supervised learning approaches to tackle the task of MCQ item difficulty estimation. We especially focused on the evaluation of model certainty as a proxy for item difficulty, but alternative approaches should also be explored. As an example, future research can focus on the impersonation ability of LLMs, by evaluating whether system instructions (e.g., "Answer this question as if you were a student that only studied half the book") can offer useful insights for difficulty estimation.

\section*{Limitations}
The main limitation of the current study is its findings' generalisability to other datasets. Specifically, as newly developed LLMs  perform increasingly well in factual knowledge benchmarks \citep{zhao2023survey}, it is expected that the model certainty metrics explored in this study will align less with student performance data (as already observed with the Llama3-70b model). Similarly, the current approach will likely also not generalise for e.g., primary-school exams, where the factual knowledge being evaluated is often trivial for an LLM.

Furthermore, in this study we observed that the question type is a significant factor to consider. However, within the currently defined question types, there exists high variability in some categories. This is mostly true for the Wh-questions, where there is a subtype of "Which statement is correct?" questions (e.g., "Which statement is true with regard to brain development?"). In total, there were 68 questions of this type, but preliminary research did not suggest that this question type differs significantly from other Wh-questions.

Generally, MCQs bear other characteristics that can also have a confounding effect. For example, we can consider the lexical (e.g., use of technical terminology) and semantic (e.g., topic) variability. However, it is not directly evident how to study this, as there is often no clear demarcation line between what constitutes, for example, a technical and a non-technical term.

Lastly \citet{tjuatja2024llms} have found that models which have undergone reinforcement learning from human feedback (RLHF) reflect even more poorly human-like behaviour. While this might be more relevant in their study than in ours, since they focus on opinions rather than factual knowledge, it might be the case that RLHF does affect correlation with human behaviour in our study, too.

\section*{Ethics Statement}
In this study, we used a dataset of multiple-choice questions from the "Biopsychology" course at the Social Sciences Faculty of the University of Groningen. The data was aggregated across multiple students and anonymised, ensuring that individual student performance cannot be traced. Because of its format as provided to us, no further ethics approval was required from our institution to work with this dataset.

\section*{Acknowledgements}
We would like to thank the Ubbo Emnius Foundation for the M20 Fund and the Jantina Tammes School for Digital Society, Technology and Artificial Intelligence for facilitating this research. We are also grateful for the contribution of Dr. Mark Nieuwenstein from the University of Groningen who provided us with the Biopsychology question set from his course. We are also thankful to the reviewers for taking the time to read and comment our paper.

\bibliography{custom,anthology}

\appendix

\section{Effect of Instruction Phrasing}
\label{app:prompt_phrasing_results}

\begin{table*}[ht]
    \centering
    \begin{tabular}{m{0.25\textwidth}  m{0.65\textwidth}}
        \toprule
        \textbf{Instruction Phrasing} & \textbf{Full Prompt} \\\midrule
        \centering{Phrasing 1} & Below is a multiple-choice question. Choose the letter which best answers the question. Keep your response as brief as possible; just state the letter corresponding to your answer with no explanation. \newline
        Question:  \newline \textit{[Question Text]} \newline Response:\\\hline

        \centering{Phrasing 2} &  You will be presented with a multiple-choice question. Select the option letter that you believe provides the best answer to the question. Keep your response concise by simply stating the letter of your chosen answer without providing any additional explanation. \newline
        Question:  \newline \textit{[Question Text]} \newline Response:
        \\
        \bottomrule
    \end{tabular}
    \caption{Instruction phrasings used for prompt sensitivity evaluation.}
    \label{tab:instruction_phrasings}
\end{table*}

\begin{figure*}[h!]
    \centering
    \includegraphics[width=1\linewidth]{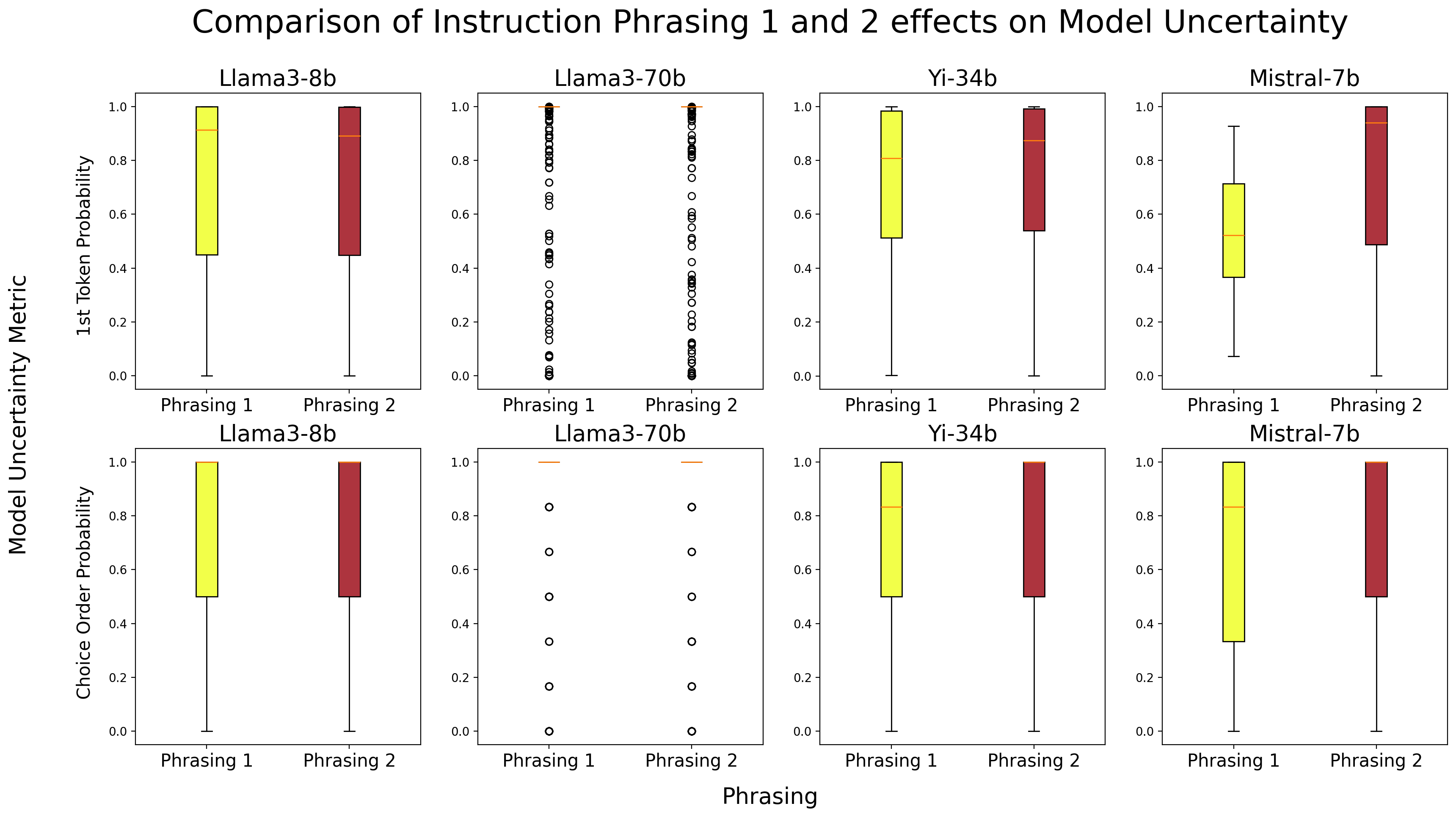}
    \caption{Effect of instruction phrasing on the two model uncertainty metrics for the selected models.}
    \label{fig:phrasing_comparison}
\end{figure*}

We study the effect of the instruction phrasing on the two model uncertainty metrics. As can be seen in Figure \ref{fig:phrasing_comparison}, a difference can be observed between the two instruction phrasings only for Mistral-7b. We further investigate whether this difference translates to considerable differences in the correlation between student and model selection rates. As can be seen in Table \ref{tab:phrasing2Mistral}, the correlation is generally slightly stronger using instruction-phrasing 1 only for Mistral-7b. For consistency, we use instruction phrasing 1 for all analyses presented in the Results section, as was also used by \citeauthor{plaut2024softmax}\citeyearpar{plaut2024softmax}.

\begin{table*}[h!]
\centering
\begin{tabular}{ccccc}
\toprule
\multirow[c]{2}{*}{\textbf{Option}} & \multicolumn{2}{c}{\textbf{1st Token Probability}} & \multicolumn{2}{c}{\textbf{Choice Order Sensitivity}} \\ \cmidrule(lr){2-3} \cmidrule(lr){4-5}
                        & Phrasing 1 & Phrasing 2 & Phrasing 1 & Phrasing 2 \\ \hline
Correct Answer & 0.229   & 0.192      & 0.252      & 0.199      \\
Distractor 1   & 0.188   & 0.113      & 0.190       & 0.116      \\
Distractor 2   & 0.134   & 0.102      & 0.132      & 0.092     \\ \bottomrule
\end{tabular}%

\caption{Effect of instruction phrasing on the observed correlation between model uncertainty and student selection rate per choice using Mistral-7b and the complete dataset.}
\label{tab:phrasing2Mistral}
\end{table*}

\section{Choice Order Sensitivity and Model Correctness}
\label{app:order_sensitivity_results}
As seen in subsection \ref{sec:question_level_chi_rates}, the two model uncertainty metrics strongly correlate, with the exception of the results from LLama3-70b. To further study this effect, we analyse the impact of choice order on the models' decisions.

Table \ref{tab:sensitivity_to_order} shows, for each model, the proportion of times the model selected the same choice across all six possible choice orderings. As can be seen, Llama3-70b is least affected by choice order (not being affected by order for 90\% of all questions), and Yi-34b is the most affected (not being affected by order for 49.4\% of all questions)

At the same time, by comparing the correctly and incorrectly answered questions subsets, we can also observe that the choice order effect is significantly smaller for correctly answered questions, showcasing that this metric is indeed a good proxy for model uncertainty.

\begin{table*}[h!]
\centering
\begin{tabular}{cccc}
\toprule
\multirow{2}{*}{\textbf{Model}} & \multicolumn{3}{c}{\textbf{Prob. of Same Choice Regardless of Order}}                \\ \cmidrule(lr){2-2} \cmidrule(lr){3-3} \cmidrule(lr){4-4}
                                & \textbf{All Questions} & \textbf{Correctly Answered} & \textbf{Incorrectly Answered} \\ \hline
Llama3-8b & 0.647 & 0.751 & 0.321 \\
Llama3-70b & 0.900  & 0.938 & 0.565 \\
Yi-34b    & 0.494 & 0.568 & 0.099 \\
Mistral-7b & 0.523 & 0.631 & 0.220  \\ \bottomrule
\end{tabular}%

\caption{Probability of selecting the same choice for all possible choice orders.}
\label{tab:sensitivity_to_order}
\end{table*}

\end{document}